\documentclass[letterpaper]{article}

\usepackage[top=30truemm,bottom=30truemm,left=25truemm,right=25truemm]{geometry}
\usepackage[square,sort,comma,numbers]{natbib}
\usepackage{mathtools}
\usepackage{here}
\usepackage{authblk}
\usepackage{setspace}
\providecommand{\keywords}[1]{\textbf{\textit{Keywords:}} #1}

\title{Neural Autopoiesis: Organizing Self-Boundary by Stimulus Avoidance in Biological and Artificial Neural Networks}
\date{\vspace{-5ex}}

\author[1]{Atsushi Masumori\thanks{Corresponding author: masumori@sacral.c.u-tokyo.ac.jp}}
\author[2]{Lana Sinapayen}
\author[1]{Norihiro Maruyama}
\author[2]{Takeshi Mita}
\author[2]{Douglas Bakkum}
\author[2,3]{Urs Frey}
\author[1]{Hirokazu Takahashi}
\author[1]{Takashi Ikegami}

\affil[1]{The University of Tokyo, Tokyo, Japan}
\affil[2]{Sony Computer Science Laboratories, Inc., Tokyo, Japan}
\affil[3]{ETH Z\"urich, Basel, Switzerland}
\affil[4]{MaxWell Biosystems AG, Basel, Switzerland}

\begin{document}
\maketitle

\begin{abstract}
Living organisms must actively maintain themselves in order to continue existing. Autopoiesis is a key concept in the study of living organisms, where the boundaries of the organism is not static by dynamically regulated by the system itself. To study the autonomous regulation of self-boundary, we focuses on neural homeodynamic responses to environmental changes using both biological and artificial neural networks. Previous studies showed that embodied cultured neural networks and spiking neural networks with spike-timing dependent plasticity (STDP) learn an action as they avoid stimulation from outside. In this paper, as a result of our experiments using embodied cultured neurons, we find that there is also a second property allowing the network to avoid stimulation: if the agent cannot learn an action to avoid the external stimuli, it tends to decrease the stimulus-evoked spikes, as if to ignore the uncontrollable-input. We also show such a behavior is reproduced by spiking neural networks with asymmetric STDP. We consider that these properties are regarded as autonomous regulation of self and non-self for the network, in which a controllable-neuron is regarded as self, and an uncontrollable-neuron is regarded as non-self. Finally, we introduce neural autopoiesis by proposing the principle of stimulus avoidance.
\end{abstract}


\section{Introduction}
Learning is an important aspect of neural systems, and it is crucial for animals, as embodied neural systems, to learn adaptive behavior to survive. One of the key concepts in the study of adaptive behavior is homeostasis. Ashby argued that adaptive behavior is an outcome of the homeostatic property of living systems \cite{Ashby1960}. Ikegami and Suzuki have proposed the concept of homeodynamics, where an autonomous self-moving state emerges from a homeostatic state \cite{Ikegami2008}, and Iizuka and Di Paolo have reported that adaptive behavior is an indispensable outcome of homeostatic neural dynamics \cite{Iizuka2007, DiPaolo2008}. However, those models are still too abstract to compare with the homeostatic property of biological neural dynamics. In this paper, we study neural homeodynamic responses to environmental changes using both biological neural cells and a more biologically plausible computational model of neurons.




Biological neural cells cultured {\it in vitro} have been used to study neural systems \cite{Potter2001, Eytan2006a, Madhavan2007, Bakkum2008} because such cultured neurons are easier to study than neurons {\it in vivo}, since they are composed of a fewer neurons, and cultured in a more stable environment. Using cultured neurons is also advantageous because unknown, complex features in neural cells can potentially be used that are still difficult to implement in artificial neural networks. Although cultured neural systems are much simpler than real brains, they have the essential properties, including spontaneous activity, various types and distributions of cells, high connectivity, and rich and complex controllability \cite{Canepari1997}. Homeostatic adaptivity may be one such property. 

Artificial neural networks are often used as models of biological neural networks to understand learning mechanisms \cite{Sejnowski1988}. Recently, the simulation of the more realistic artificial neural networks have become computationally efficient through the introduction of models of spiking neurons \cite{Maass1997, Izhikevich2004, Brette2007} and of synaptic plasticity \cite{Song2000, Dan2006, Caporale2008}. These more realistic models can lead to theoretical understanding of biological neural networks.

Following the organization of biological experiments and the development of a computational model, we propose the homeodynamic control of a neural network, based on the new learning principle of synaptic plasticity, by which neural networks learn to avoid stimulation from the environment.

\subsection{Learning by Stimulation Avoidance}
Shahaf and Marom demonstrated that a cultured neural network can learn a desired behavior as if the network avoid stimulation using the following protocols \cite{Shahaf2001}. First, an electrical stimulation with a fixed low frequency (e.g., 1--2Hz) was delivered to a predefined input zone of the network. When the desired behavior appeared, the stimulation was removed. After this protocol was repeated, the network learned to produce the expected behavior in response to the stimulation. In practice, the authors showed that the networks learned to produce spikes at predefined output zones, in a predefined time window (40--60 ms after each stimuli) in response to the stimulation applied at the input zone. 

Marom and Shahaf explained these results by invoking the stimulus regulation principle (SRP) \cite{Marom2002}. SRP is composed of the following two functions at the neural network level: (i) modifiability, in which stimulation drives the network to try to form different topologies by modifying neuronal connections, and (ii) stability, in which removing the stimulus stabilizes the last configuration of the network. They argued that their preliminary experiments suggested that cultured neural networks had these two functions at the network \cite{Shahaf2001}. However, the two properties which constitute SRP are macroscopic phenomenological explanations, which do not form a concrete mechanism. Furthermore, assuming that modifiability is correct, learned configurations are destroyed at every stimulation; if stability is correct, it should not be necessary to repeat such cycle as in the experiment explained above.

In a previous study, we proposed a mechanism at the micro scale of neural dynamics using small simulated networks to explains Shahaf and Marom's results, termed Learning by Stimulation Avoidance (LSA) \cite{Sinapayen2016}. LSA is based simply on a classical form of spike-timing dependent plasticity (STDP) \cite{Song2000}. When a presynaptic neuron and a postsynaptic neuron fire successively within a certain time window, the synaptic weight increases (long-term potentiation, or LTP). If the timing of the two neurons' firing is reversed, the weight decreases (long-term depression, or LTD). STDP has been found in both {\it in vivo} and {\it in vitro} networks \cite{Caporale2008}.
 
The following two dynamics emerge for the avoidance of stimulation in LSA, based on STDP. The first dynamics is the reinforcement of behavior that decreases the stimulation by LTP. If the firing of the postsynaptic neuron terminates the cause of presynaptic neuron, then the synaptic weight from presynaptic neuron to the postsynaptic neuron will increase. Thus the behavior leading to the decrease in stimulation is reinforced. The second dynamics is the weakening of behavior that increases the stimulation by LTD. If the firing of the postsynaptic neuron initiates the cause of presynaptic neuron, then the synaptic weight from the presynaptic neuron to the postsynaptic neuron will decrease. Thus the behavior leading to the increase in stimulation is weakened. This is the most basic structure of LSA.

LSA states that the network learns to avoid external stimulus by learning available behaviors. In the follow-up studies, we further showed that LSA scales up to larger networks \cite{Masumori2017a}. LSA works, as long as the following two conditions are met: (i) the plasticity of the embodied neural network is driven by STDP, and (ii) the network constitutes a closed-loop with the environment. We claim that LSA is an emergent property of spiking networks with Hebbian rules \cite{Hebb1949} and its environment.
 
Such stimulation avoidance can produce homeostasis: since the unexpected stimulation of an agent resulting from its environment represents environmental changes, avoiding the stimulation decreases the influence of environmental changes on the internal state of the agent. It is interesting that such homeostatic global behaviors directly emerges from simple synaptic dynamics, such as STDP. This property of stimulus avoidance can also be regarded as an intrinsic motivation that emerged from the local dynamics of neurons in a bottom-up manner. 


\subsection{Motivation}
Shahaf and Marom's results are promising in that they showed that behaviors that avoid stimulation will be learned autonomously in cultured neurons \cite{Shahaf2001}. However, there have been no further studies on this learning dynamics. Although Shahaf and Marom used cultures with 10,000--50,000 neurons, a smaller number of cultured neurons would be able to learn in the same manner, if the same dynamics as LSA works in cultured neurons. Thus we first performed learning experiments using a smaller number of cultured neurons than previous works have used, to demonstrate how such a learning mechanism scales from small to large cultured neurons. 

In addition, the results of the experiments show that if the network cannot learn a behavior that removes external stimuli, its response to stimulation is gradually suppressed, as if it isolates the uncontrollable input neurons. This means that the second property, where the network avoids the effects of stimulation through the weakening the connection from uncontrollable input neurons, is used alongside LSA to avoid stimulation.

We also find such a behavior can be reproduced by simulated spiking neural networks with asymmetric STDP, where the functions of LTP and LTD are rotationally asymmetric (e.g., the working window for LTD is larger than the window for LTP). These dynamics can be interpreted as that the embodied neural network autonomously regulates the boundary between self and non-self. 

Furthermore, it should be noted here that updating the concept of autopoiesis was a broad motivation for this study. 
Autopoiesis has been proposed by \cite{Maturana1980} as a fundamental principle of living systems. A microscopic level of a system's organization is iteratively compensated for by the emerging macro state; taking cell dynamics as an example, as the resultant organization is a cellular boundary organized in a chemical space that distinguishes between self and non-self. This perspective is commonly applied to both the immune and neural systems of vertebrates. Immune systems can distinguish between the material self and non-self and neural systems can distinguish between ``informational'' self and non-self. 
The neural autopoiesis demonstrated in this paper is based on the principle of stimulus avoidance observed in neural networks. 

\section{Materials and Methods}
We used both cultured neurons and spiking neural networks for studying the homeostatic properties in embodied neural networks. Below, we first describe the methods of cultured neurons, then the methods of spiking neural networks and each experimental setup. 

\subsection{Cultured Neurons}
\subsubsection{Cell Culture}
The neural cultures were prepared from the cerebral cortex of E18 Wistar rats, as previously reported \cite{Bakkum2013, Yada2016, Yada2017}. The cortex region was trypsinized with 0.25\% trypsin, and the dissociated cells were plated and cultured on a recording device. The surfaces of the electrodes on the device were coated with 0.05\% polyethylenimine and laminin to improve plating efficiency. The cells were cultured in Neurobasal Medium (Life Technologies, California, United States) containing 10\% L-Glutamine (Life Technologies, California, United States) and 2\% B27 supplement (Life Technologies, California, United States) for the first 24 h. After the first 24 h, half the plating medium was replaced with growth medium (Dulbeccos modified Eagles medium (Life Technologies, California, United States)) that contained 10\% horse serum, 0.5 mM GlutaMAX (Life Technologies, California, United States), and 1 mM sodium pyruvate. The cultures were placed in an incubator at 37$^\circ$C with an H$_2$O-saturated atmosphere that consisted of 95\% air and 5\% CO$_2$. During cell culturing, half the medium was replaced once a week with the growth medium. The all cultures used in our experiments consisted of 47--110 cells and were sufficiently matured to show global burst synchronization. 

\subsubsection{CMOS-based High-Density Microectrode Arrays}
\begin{figure}[htbp]
\centering
\includegraphics[width=12cm]{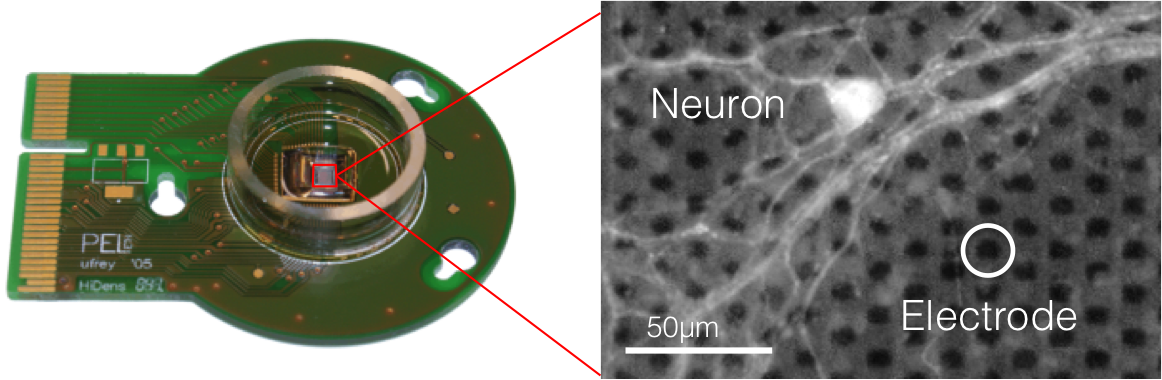}
\caption{The high-density CMOS electrode array used in this experiment. This recording device has 11,011 recording sites, a diameter of 7 $\mu$m, and an inter-electrode distance of 18 $\mu$m.}
\label{fig:cmos}
\end{figure}

A high-density microelectrode array based on complementary metal-oxide semiconductor (CMOS) technology \cite{Frey2010} was used to measure the extracellular electrophysiological activity of the cultured neurons (Fig.~\ref{fig:cmos}). This CMOS-based electrode array is superior to the conventional multi-electrode array (MEA) used previously \cite{Potter2001}, in that it has a far higher spatio-temporal resolution. The number of electrodes in conventional MEAs is small (e.g., 64 electrodes), and the locations of the recording electrodes are predetermined using a large inter-electrode distance (e.g., 200~$\mu$m). Thus, it is difficult to identify signals from an individual cell. In contrast, the CMOS arrays have 11,011 electrodes; the diameter of the electrode is 7 $\mu$m, with an inter-electrode distance of 18 $\mu$m over an area of 1.8 mm $\times$ 1.8 mm. Thus this method can identify signals from an individual cell in a small culture. The device can simultaneously record the electrical activity of 126 electrodes at a sampling rate of 20 kHz.

\subsubsection{Estimation of Neuronal Somata Locations}
Before recording the neural activities, the 11,011 electrodes were scanned to obtain an electrical activity mapping to estimate the locations of the neuronal somata (i.e., identifying the positions of neural cells). The scanning session consisted of 95 recordings. In each recording, the electrical activities for 110--120 electrodes were simultaneously recorded for 60 s. In the recordings, the sampling frequency was set to 20 kHz and the band path filter was set to 0.5--20 kHz. An electrical activity map was obtained by averaging the height of the action potentials for each electrode. We applied a Gaussian filter to the map and assumed that the neuronal somata were located near the local peaks on the Gaussian-filtered map. At most, 126 peaks were selected, in descending order, as the positions of neural cells, and the electrodes nearest the peaks were used to record the neural activity. If the number of local peaks were fewer than 126, then all the peaks were used. Using this method, one electrode can ideally represent a single neural state. 

\subsubsection{Estimation of Excitatory and Inhibitory Synaptic Conductances}
It was important that we were able to identify the neuron's type for these experiments, as the neural activity were recorded, and stimulation was applied, for each neuron, not for a group of neurons. The neuronal cell type (i.e., excitatory or inhibitory) was estimated using spike shapes, which were recorded for 10 min before the main experiment \cite{Tajima2017}. The shapes of the action potential of these two neural types differ: in the action potential of excitatory neurons, the distance between a maximum potential and a minimum potential is longer than that of inhibitory neurons. We classified the type of neuronal cell by using k-means clustering \cite{macqueen1967} according to the difference in shapes. Here, we used the average length between the two peaks of the spike shape to classify those groups into two classes ($k=2$).

\subsubsection{Recording and Preprocessing of Neural Activity}
To detect and record the spike in cultured neurons, we used MEABench software developed by \cite{Wagenaar2005b}. All recordings were performed at a 20~kHz sampling rate using the real-time spike detection algorithm LimAda in the MEABench. As the LimAda algorithm detects spikes that exceed the threshold, but without distinction between positive and negative values, unexpected double detection of spikes can occur. These unexpected double-detected spikes were removed from the data before analysis. By sending the electrical stimuli to a neuronal cell through the electrodes, artifacts might occur. In our experiments, we need to detect the action potential and stimulate the neuronal cell at the same time. The Salpa filter in MEABench was used to remove the artifacts in real time \cite{Wagenaar2002}.

\subsection{Spiking Neural Networks}
\subsubsection{Neuron Model}
The model for spiking neurons proposed by \cite{Izhikevich2004} was used to simulate excitatory neurons and inhibitory neurons. This model is well known, as it can be regulated to reproduce the dynamics of many variations of cortical neurons, and it is computationally efficient. The equations of the neural model are defined as follows: 

\begin{eqnarray}
\frac{dv}{dt} = 0.04v^2 + 5v + 140 -u + I,\\
\frac{du}{dt} = a(bv - u),\\
\text{if } v \geq 30 \text{mV},~\text{then}
\begin{cases}
    v \leftarrow c\\
    u \leftarrow u+d.
\end{cases}
\end{eqnarray}

\begin{figure}[htbp]
\centering
\includegraphics[width=14cm]{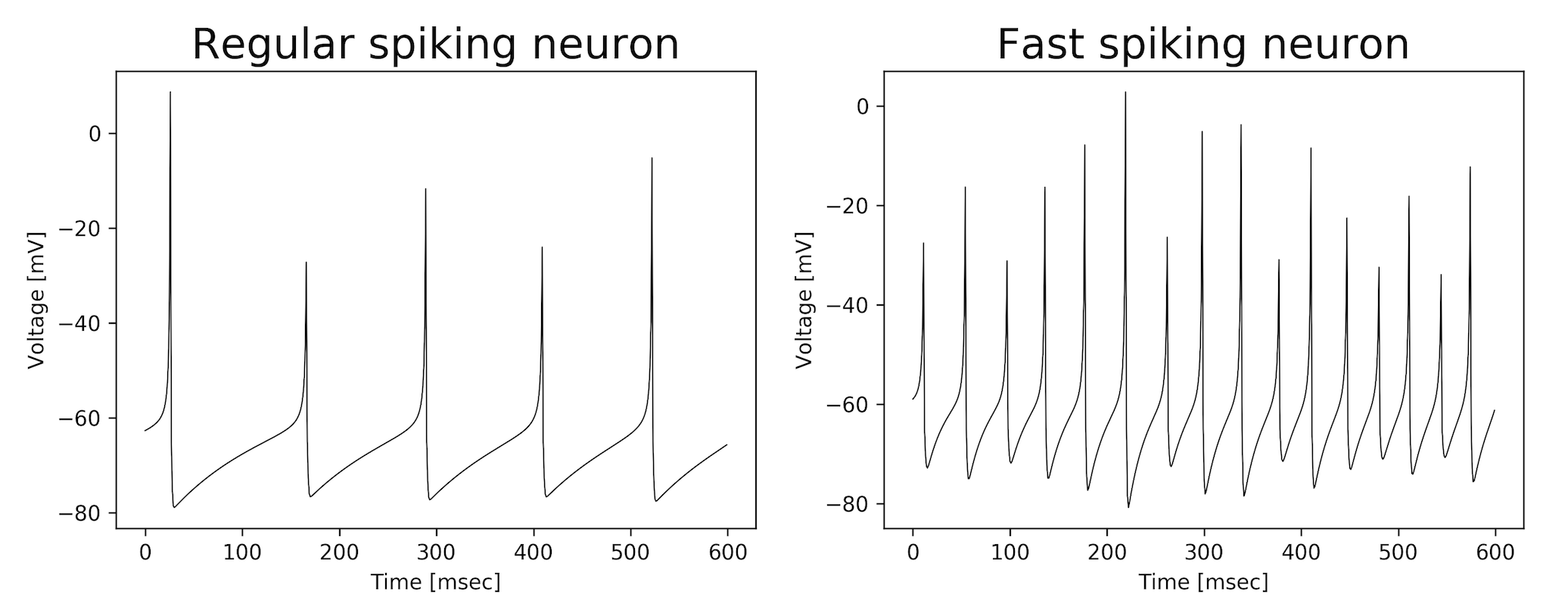}
\caption{Dynamics of regular-spiking and fast-spiking neurons simulated using the Izhikevich model. Regular-spiking neurons were used as excitatory neurons and fast-spiking neurons were used as inhibitory neurons.}
\label{fig:IzhDynamics}
\end{figure}

Here, $v$ represents the membrane potential of the neuron, $u$ represents a variable related to the repolarization of membrane, $I$ represents the input current from outside of the neuron, $t$ is the time, and $a, b, c$, and $d$ are other parameters which control the shape of the spike \cite{Izhikevich2003a}. The neuron is regarded as firing when the membrane potential $v$ exceeds 30~mV. The parameters for excitatory neurons (regular-spiking neuron) were set as: $a = 0.02,~b = 0.2,~c = -65~\text{mV}$, and $~d = 8$, and for inhibitory neurons (fast-spiking neuron) are set as: $a = 0.1,~b = 0.2,~c = -65~\text{mV}$, and $~d = 2$ (Fig.~\ref{fig:IzhDynamics}). The simulation time step $\Delta t$ is 0.5~ms.
The parameter values were chosen to reflect biological relevance \cite{Izhikevich2003a}.

The input current $I$ were added for each neuron $n_i$ at each time step $t$ as:

\begin{equation}
\label{eq:neuronInput}
\begin{split}
&I_i = I_i^*+ e_i + m_i\\
&I_i^*= \sum_{j=0}^{n}f_jw_{ji}s_j \;\\
&f_j =  
\begin{cases}
    1 , & \text{if neuron $j$ is firing}\\
    0, & \text{otherwise}.
\end{cases}
\end{split}
\end{equation}

Here, $m$ represents zero-mean Gaussian noise with standard deviation $\sigma=3$~mV delivered to each neuron at each time step as internal noise input; $e$ represents external stimulation (conditions, frequency and strength of the external stimulation are described in a later section); $s$ represents short-term plasticity variables. A phenomenological model of short-term plasticity (STP) \cite{Mongillo2008} was used here. STP is a reversible plasticity rule that decreases the intensity of neuronal spikes if they are too close in time. As in the original paper, we applied STP to the output weights from excitatory neurons to both excitatory and inhibitory neurons. Practically, $s$ varies for each neuron $n_j$ as:

\begin{equation}
\label{eq:stp}
\begin{split}
&s_j = u_jx_j\\
&\frac{dx}{dt} = \frac{1-x_j}{\tau _d} -u_j~x_j~f_j\\
&\frac{du}{dt} = \frac{U-u_j}{\tau _f} +U(1-u_j)~f_j.\\
\end{split}
\end{equation}

Here, $x$ represents the amount of available resources and $u$ represents the resource used by each spike \cite{Mongillo2008}. The parameters were set to $\tau_d = 200$~ms, $\tau_f=600$~ms, and $U=0.2$~mV.

STP is not necessarily required for LSA, but it is efficient to suppress strong global burst synchronization in an initialization phase of spiking neural networks \cite{Sinapayen2016} and stabilizes the firing rate \cite{Masumori2018}. LSA and the second property of stimulation avoidance studied in this paper can be achieved only by a parameter tuning without STP (e.g., strength of noise input, initial values of weights, or learning rate of STDP). However, it can be easily achieved with STP (e.g., results of experiments without STP can show the almost same tendency with the results in this paper, but require more simulation time). Moreover, applying STP also makes the networks more realistic. We thus used the STP model in this study. 

\subsubsection{Spike-Timing Dependent Plasticity}
A computational model of the classical form of STDP was used as a model for synaptic plasticity. It updates the synaptic weight between two connected neurons depending on the relative timing of their spikes; when the presynaptic neuron fires immediately before the postsynaptic neuron, the synaptic weight increases, and, in the opposite case, the synaptic weight decreases. The amount of weight change $\Delta w$ obeys the following dynamics:

\begin{equation}
\label{eq:stdp}
\Delta w =  
\begin{cases}
    A_{LTP}(1-\frac{1}{\tau_{LTP}})^{\Delta t} , &\text{if } \Delta t > 0\\
    -A_{LTD}(1-\frac{1}{\tau_{LTD}})^{-\Delta t} , &\text{if } \Delta t < 0.\\
\end{cases}
\end{equation}

\begin{figure}[htbp]
\centering
\includegraphics[width=12cm]{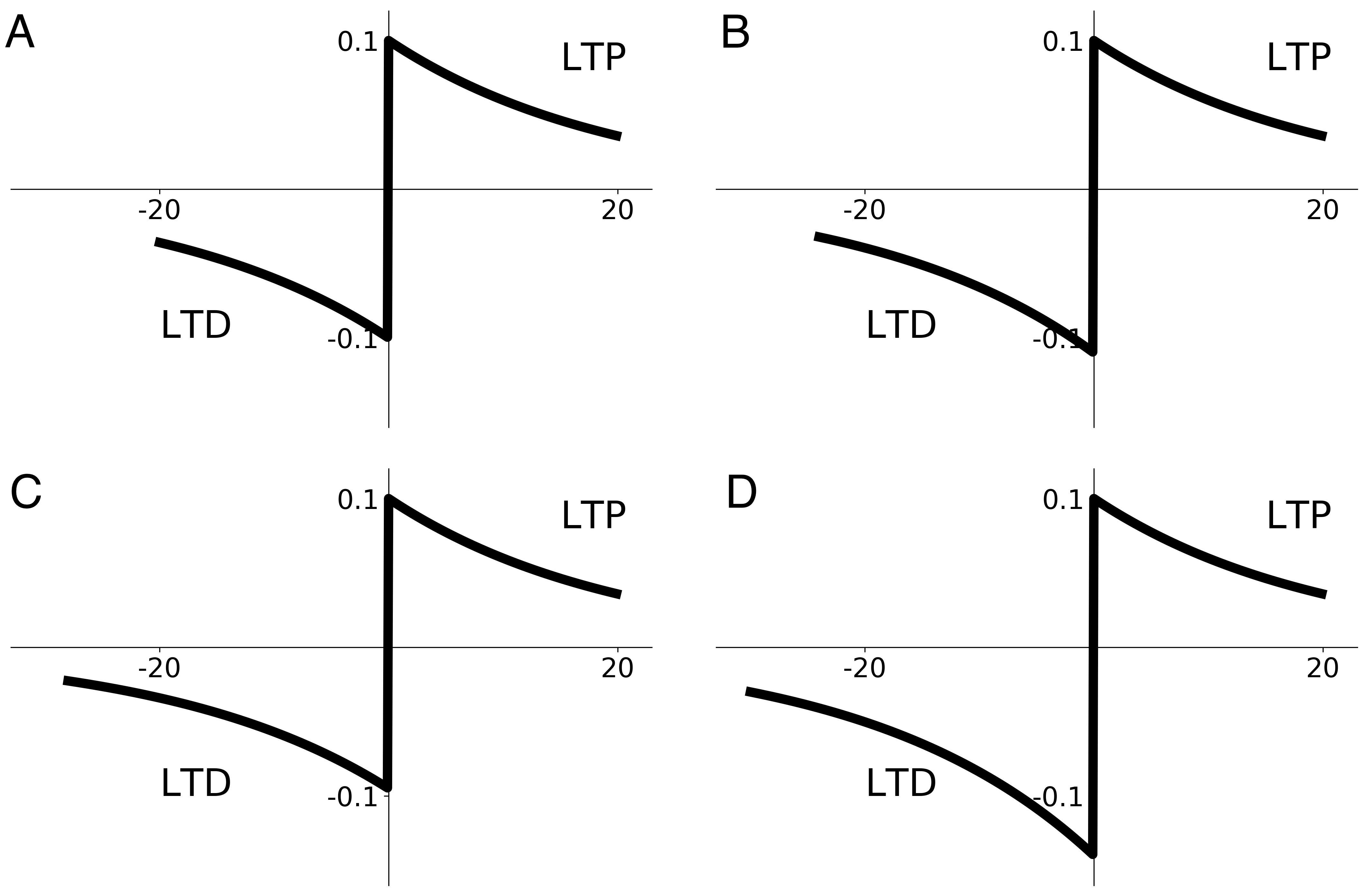}
\caption{Parametric variations of the STDP curve. A: Symmetric STDP: $A_{LTP}$, $A_{LTD}=1.0$; $\tau_{LTP}$, $\tau_{LTD}=20$. B: Example of asymmetric STDP: $A_{LTP}=1.0$, $A_{LTD}=1.1$; $\tau_{LTP}=20$, $\tau_{LTD}=24$; the peak of LTD is higher and the tail of LTD is longer than that of LTP. C: Example of asymmetric STDP: $A_{LTP}=1.0$, $A_{LTD}=0.95$; $\tau_{LTP}=20$, $\tau_{LTD}=28$; the peak of LTD is smaller and the tail of LTD is longer than that of LTP. D: Example of asymmetric STDP: $A_{LTP}=1.0$, $A_{LTD}=1.4$; $\tau_{LTP}=20$, $\tau_{LTD}=30$; the peak of LTD is higher and the tail of LTD is longer than that of LTP.}
\label{fig:stdp}
\end{figure}

Here, $\Delta t$ represents the relative spike timing between the presynaptic neuron $a$ and the postsynaptic neuron $b$: $\Delta t = t_b - t_a$ ($t_a$ represents the time of the spike of presynaptic neuron $a$, and $t_b$ represent the timing of the spike of postsynaptic neuron $b$). $A_{LTP}$ and $A_{LTD}$ are the parameters for the strength of the effect of LTP and LTD in STDP. $\tau_{LTP}$ and $\tau_{LTD}$ are the parameters for the working window of LTP and LTD in STDP. The parameters $A_{LTP}, A_{LTD}, \tau_{LTP}, and \tau_{LTD}$ were varied depending on experiments, to use various forms of STDP. For symmetric STDP, the parameters of LTP and LTD that were rotationally symmetrical ($A_{LTP}$, $A_{LTD}=1.0$; $\tau_{LTP}$, $\tau_{LTP}=20$) were used; for asymmetric STDP, the dynamics of LTP and LTD that were rotationally asymmetrical ($A_{LTP}=1.0$, $A_{LTD}=0.8\text{--}1.5$; $\tau_{LTP}=20$, $\tau_{LTD}=20\text{--}30$) were used. Fig~\ref{fig:stdp} shows the variation of $\Delta w$ depending on $\Delta t$ in the symmetric STDP, along with three examples of asymmetric STDP.

STDP was applied only between excitatory neurons; thus, the weight of other connections did not change from their initial value for all experiments. The weight value $w$ between excitatory neurons varies as:

\begin{eqnarray}
w_t = w_{t-1}+ \Delta w .\;
\label{eq:wvariation}
\end{eqnarray}

The maximum possible weight is fixed to $w_{max}=20$, and if $w > w_{max}$, then $w$ is reset to $w_{max}$. The minimum possible weight is fixed to $w_{min}=0$, and if $w < w_{min}$, then $w$ is reset to $w_{min}$. 

In addition to STDP, a weight decay function was applied to the weights between excitatory neurons. The decay function is defined as follows:

\begin{eqnarray}
w_{t+1} = (1 - \mu) w_t.
\label{eq:decay}
\end{eqnarray}
 
The parameter $\mu$ was fixed as $\mu = 5 \times 10^{-7}$.

\subsection{Experimental Setup}
We performed learning experiments using the neuronal cultures with the same settings as before \cite{Masumori2015, Masumori2018}. 
Then we repeated the same experiments with the simulation model using the spiking neural network and examined whether the results from the experiments using real neurons could be reproduced by the model. 

In the experiments with neuronal cultures, we performed two types of experiments. Experiment 1 used a robot in one-dimensional virtual space (Fig.~\ref{fig:virtual}); experiment 2 used a robot in two-dimensional real space (Fig.~\ref{fig:realspace}). In both cases, the stimulation was applied as the sensor input when the robot approached the wall, and the sensor input stopped when the robot moved away from the wall. Those input neurons were randomly selected from a population. We explain each experimental setup in detail in following sections.

\subsubsection{Embodied Cultured Neurons in One-Dimensional Virtual Space}
In experiment 1, a virtual robot was coupled to the neuronal culture via the CMOS arrays, which could detect neural activity and inject electrical stimuli as explained above. Input and output neurons were determined in the following way. Input neurons were randomly chosen from excitatory neurons. The number of input neurons depended on each experiment (2 or 10 neurons). Before starting the learning experiments, 20 stimuli at 1~Hz were applied to the input neurons and the neural activity was recorded. Based on the recorded data, 10 output neurons were chosen from excitatory neurons to satisfy the following requirement: following the 20 stimuli, the mean number of spiked neurons within 20 to 40 ms following each stimulus is less than 5. With these input and output neurons, the network cannot behave to avoid stimulation in the initial state. This procedure was performed using 10 selected input neurons, and if there was no combination of such output neurons, then this procedure was performed using the 2 selected input neurons.

The virtual robot moved forward at a constant speed; if the robot approached a wall, the sensors stimulated the input neurons and if more than 5 out of 10 output neurons fired within the specific time window following the stimulation, then the robot would turn away from the wall, rotating at 180 degrees \cite{Masumori2018}. We call this setup  the closed-loop condition. This cycle was repeated 10 times per experiment. We performed six experiments with three cultures (10--43 days {\it in vivo}) with the settings. We also performed an experiment with an open-loop condition, where the stimulus input stopped at random, regardless of the network's neural activity (the other settings were same as under the closed-loop conditions).

\subsubsection{Embodied Cultured Neurons in Two-Dimensional Real Space}
\begin{figure}[htbp]
\centering
\includegraphics[width=12cm]{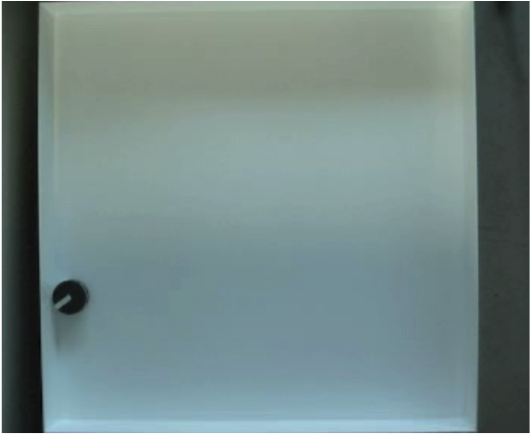}
\caption{Experimental environment of the robot experiment in two-dimensional real space. The robot was placed in a flat square arena (60 cm $\times$ 60 cm).}
\label{fig:realspace}
\end{figure}

In experiment 2, Elisa-3 (GCtronic, Ticino, Switzerland) was used as the mobile robot in real space (Fig.~\ref{fig:realspace}). Elisa-3 is a small, circular robot with a 2.5~cm radius and has two independently controllable wheels. The front right and front left distance sensors were used as sensory signals to stimulate the neuronal cells. The refresh rate of the robot was 10 frames per second.

A simple sensorimotor mapping was applied to the robot and the cultured neurons. We randomly selected 2 neurons, estimated to be excitatory neurons, as the left- and right-input neurons. At given time intervals (100 ms), the probability $P_{L,R}$ to send electrical stimulation to the input neuron was controlled by the sensory value of the mobile robot. Specifically, the probability was calculated as:

\begin{eqnarray}
  P_{L,R} = \left\{ \begin{array}{ll}
    0 & S_{L,R} < T \\
    S_{L,R}/S_{max} & S_{L,R} \geq T.
  \end{array} \right.
\end{eqnarray}
 
If sensor value $S_{L,R}$ (0--950) was less than threshold $T$, then $P_{L,R}$ became zero. Otherwise $P_{L,R}$ was calculated using $S_{L,R}/S_{max}$. $S_{max}$ denotes a maximum value of the sensor input (950). Whether the stimulus will be delivered to the input neuron or not was determined using this probability every 100 ms. The threshold $T$ was set to 100. According to this form, the distance from the robot to the wall was encoded as the stimulation frequency.
 
We also randomly selected 20 neurons, estimated to be excitatory neurons, as output neurons: 10 neurons were used for calculating the speeds of each left and right wheels respectively. The wheel speeds were calculated based on the number of spikes of the output neurons that were integrated every 100 ms. The left and right wheel speeds $V_{L,R}$ were calculated as:

\begin{eqnarray}
V_{L,R} = k\sum_{i\in {N}_{L,R}} v_i + C_{L,R}
\end{eqnarray}

The positive integers $v_i$ were equal to the number of spikes of the output neurons over a given time interval (100 ms), and sum them with the negative constant weight $k$, and a positive constant $C$ as a default wheel speed was added. ${N}_L$ and ${N}_R$ were the sets of left- and right-output neurons. Here, as $k$ was a negative value and $C$ is a positive value, the robot moved forward when the output neurons are not active. $k$ was set to -0.3. The default values of $C_{L,R}$ were 12.5 and the values were adjusted respectively before the experiments so the robot would go straight without the firings of output neurons. As the activity of the output neurons increased, the speed of the forward movement decreased, until, finally, the robot moved backwards. Since the two wheels of the robot were independent, the robot could turn when the wheel speeds were different. 

Because of these settings, as the robot approached the wall, the sensor values became higher and the stimulus input was applied. The number of spikes of the output neurons was the coefficient of the left and right motor speeds of the robot; thus to avoid the wall, the robot needed to control the motor speeds. 

These conditions were more difficult for the network to learn an action to avoid the wall, compared with experiment 1. We performed six experiments with three cultures (26--61 days in vivo, but a technical issue occurred in one of the experiments, so the recorded data for the other five experiments were analyzed.)


\subsubsection{Embodied Spiking Neural Networks in One-Dimensional Virtual Space}
We also performed a simulation experiment similar to experiment 1, which used the simulated spiking neural network. The model for spiking neurons, proposed by Izhikevich \cite{Izhikevich2004} and explained above, was used to simulate excitatory neurons and inhibitory neurons. The network consisted of 80 excitatory neurons and 20 inhibitory neurons. This ratio of inhibitory neurons is standard in simulations \cite{Izhikevich2003a, Izhikevich2004} and similar to biological values \cite{Cassenaer2007}. The excitatory neurons were divided into three groups: input (10 neurons), output (10 neurons), hidden (60 neurons). The networks were fully connected; the weight values $w$ between each neuron were randomly initialized with uniform distributions as $0<w<5$ for excitatory neurons, $-5<w<0$ for inhibitory neurons. Only connections between excitatory neurons had synaptic plasticity based on STDP, and the weight values of other connections did not change. Both symmetric and asymmetric STDP, explained above were used for the synaptic dynamics. 

In the simulation experiment, two types of external stimulation conditions were applied: closed-loop and open-loop condition. In the closed-loop condition, stimulation was delivered at a fixed frequency (100~Hz) with 10~mV, and if more than 5 out of 10 output neurons fired within 10 ms after the stimulation, then the stimulation was removed for 1,000--2,000 ms (randomly chosen each time). Under these conditions, it was possible for the network to learn a behavior to avoid the stimulation. In the open-loop condition, the stimulation was randomly removed, regardless of the firing of output neurons, and any other settings of the stimulation were same with the closed-loop condition. In this condition, the network could not learn any behavior that avoid the stimulation.

\section{Results}
We first show the results of experiment 1 and 2, then the results of the simulation experiments. In the analysis of experiment 1 and 2, we investigated the neural dynamics in conditions where it was difficult to learn stimulus-avoidance behavior; we focus on the results of the open-loop condition in experiment 1, in which it is impossible for the network to learn that behavior, as well as the results of experiment 2, in which more networks failed to learn the behavior compared to experiment 1.

\subsection{Cultured Neurons Learn Stimulus-Avoiding Behaviors}
\subsubsection{Experiment 1}
\begin{figure}[H]
\centering
\includegraphics[width=12cm]{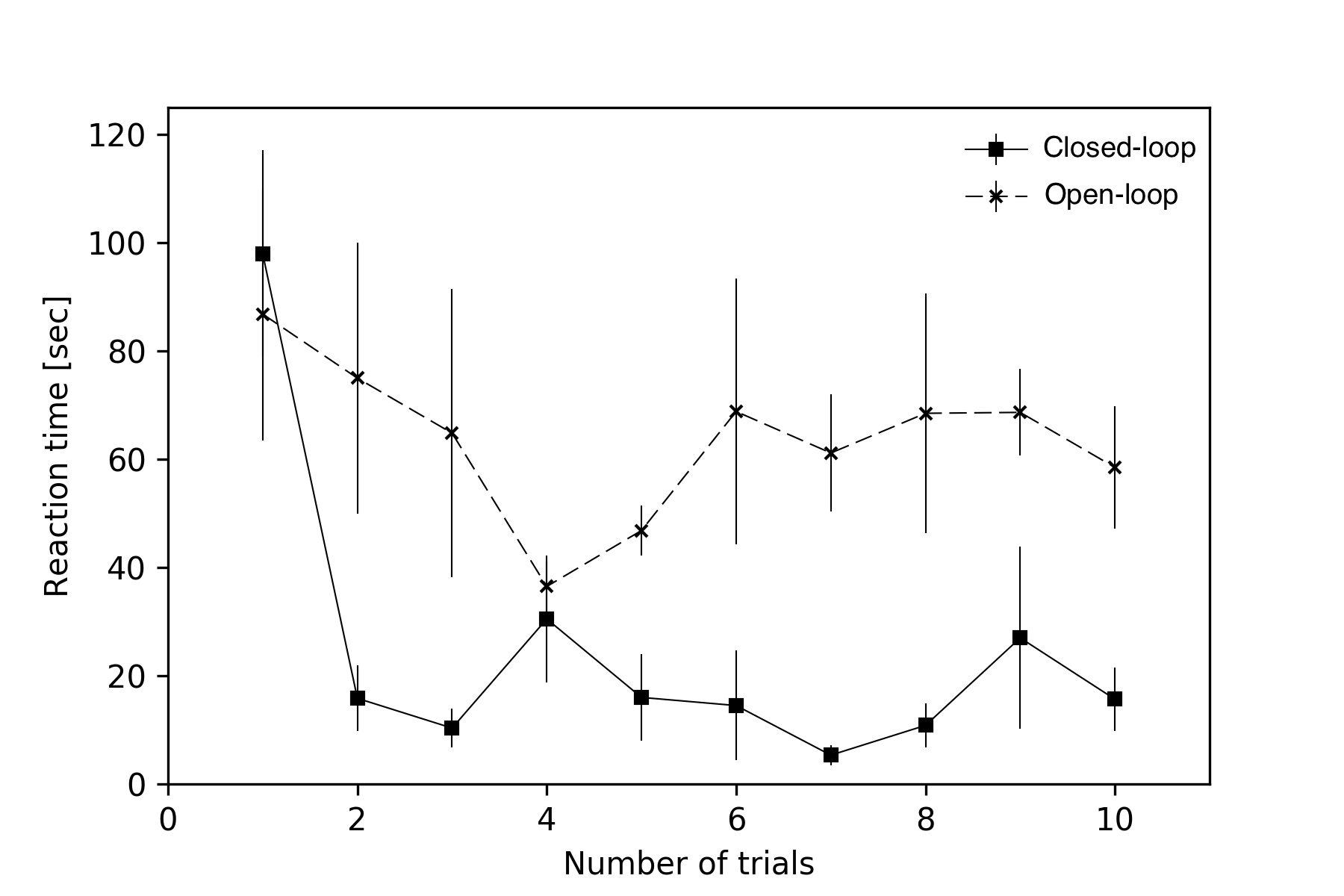}
\caption{Learning curve for both the closed-loop condition and the open-loop condition, where the stimulus is randomly applied. Statistical results for n~=~6 with standard error.}
\label{fig:learncurv}
\end{figure}

We evaluated the learning results using the reaction time (i.e., time from the beginning of the stimulation to the time of wall avoidance). Figure~\ref{fig:learncurv} shows the learning curves from experiment 1. The vertical axis represents the reaction time, where lower values indicate higher learning ability. As shown in Fig.~\ref{fig:learncurv}, in the closed-loop condition, the reaction time rapidly decreased and stabilized, indicating higher learning ability. On the other hand, in the open-loop condition, where the stimulus was randomly applied, the reaction time did not stabilize at lower values and the variance was higher than that in the closed-loop condition. 

\begin{figure}[ht]
\centering
\includegraphics[width=12.0cm]{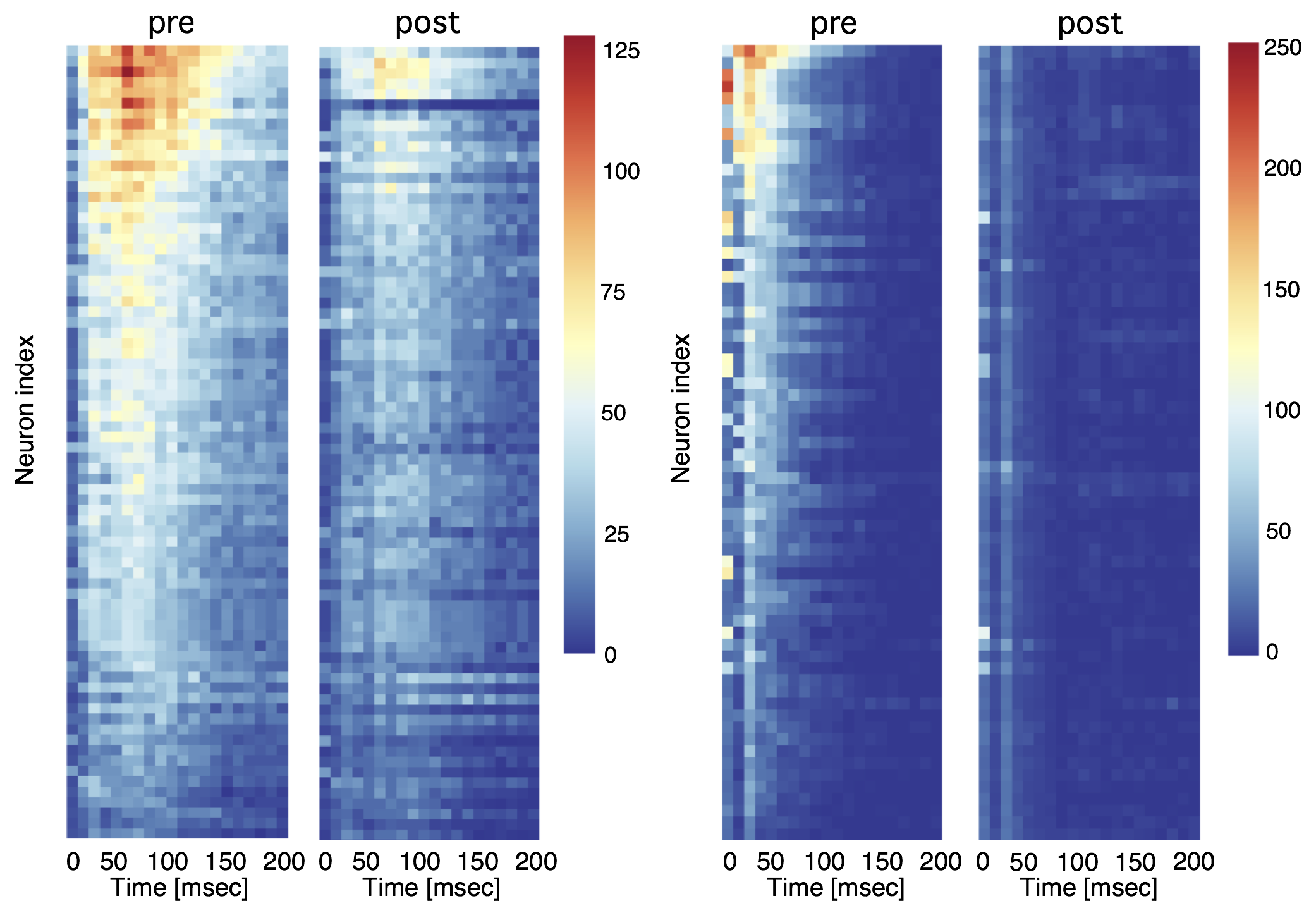}
\caption{Stimulus-evoked spikes of all neurons in the open-loop conditions where the stimulus was randomly applied. ``pre'' display the first 5 min of the experiment, and ``post'' displays the last 5 min. The evoked spikes at the end of the experiment (i.e., according to ``post'') had decreased.}
\label{fig:evoked_sample}
\end{figure}

These results are similar to the results from previous experiments involving large number of cultured neurons (10,000--50,000 neurons) \cite{Shahaf2001}. This suggest that such learning behaviors scale from small to large cultured neurons. In addition, we found that the stimulus-evoked firing considerably decreased in some cases of the open-loop conditions where the stimulus was randomly applied (Fig.~\ref{fig:evoked_sample}), although the stimulus-evoked firing increased in closed-loop conditions \cite{Masumori2018}. Below, we focus on the data from the open-loop conditions. 

\begin{figure}[htbp]
\centering
\includegraphics[width=12cm]{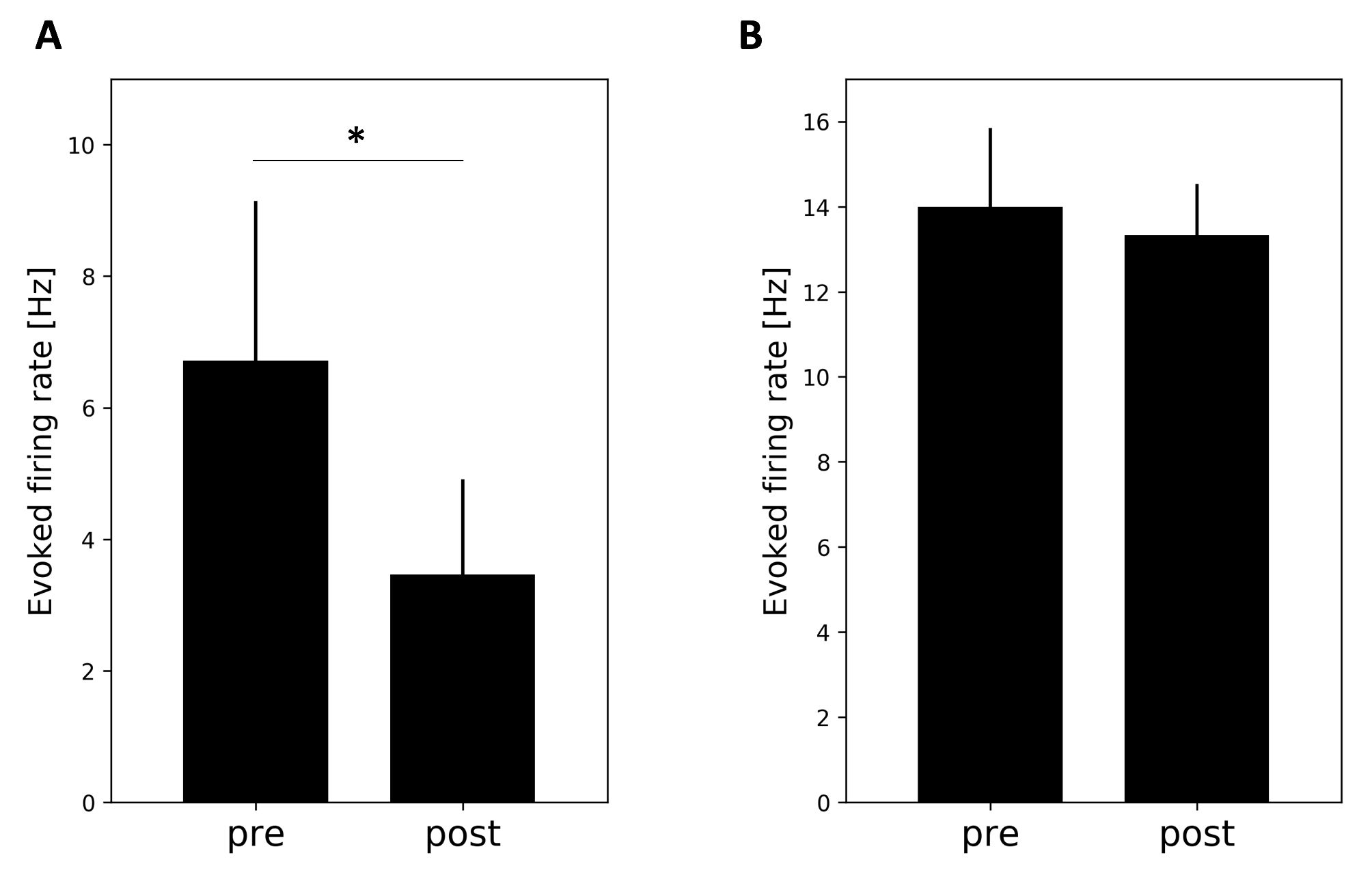}
\caption{Mean evoked firing rates with the standard errors in the open-loop conditions where the stimuli were randomly applied (n~=~6). ``pre'' is the first 5 min in the experiment; ``post'' is the last 5 min. A: mean evoked firing rates of all neurons, except for input neurons. ($\text p<0.03$). B: mean evoked firing rates of input neurons.}
\label{fig:evoked_stats_random}
\end{figure}

Figure~\ref{fig:evoked_stats_random} shows mean evoked firing rates. The evoked firing rate, except for input neurons (Fig.~\ref{fig:evoked_stats_random}A), consisted of spikes within 200~msec after each stimulus. The evoked firing rate of input neurons (Fig.~\ref{fig:evoked_stats_random}B) consisted of spikes 50~msec after each stimulus; this time window differs from the window mentioned above because we focused on the evoked spikes by each stimulus, excluding the evoked spikes by feedback from other neurons.

Besides the qualitative results in Fig.~\ref{fig:evoked_sample}, the statistical results show that the mean evoked firing rates for all neurons, except for input neurons, significantly decreased in the last 5 min of the experiments, relative to the first 5 min (Wilcoxon signed-rank test, n~=~6, p~=~0.012) (Fig.~\ref{fig:evoked_stats_random}A). On the other hand, the evoked firing rates of input neurons did not change significantly (Fig.~\ref{fig:evoked_stats_random}B). These results imply that the decrease in the number of evoked spikes was not caused by a decrease in the firing rates of the input neurons. This suggests that the cultured neurons tried to learn a behavior to avoid an external stimulation, but if the network could not avoid the stimulus, it tended to ignore the external stimulation. 

\subsubsection{Experiment 2}
In experiment 2, the learning of wall avoidance behavior succeeded in only 2 out of 5 experiments. Here, success was defined as the reaction time (i.e., time from the start of stimulation to the time of wall avoidance) decreasing by 30\% or more. The average reaction time from the first 10 min and the last 10 min of the experiments were used to calculate the success rate. We focused on dynamics where LSA failed, examining if dynamics similar to those from the open-loop conditions in experiment 1 were observed in this more difficult task.

\begin{figure}[htbp]
\centering
\includegraphics[width=15cm]{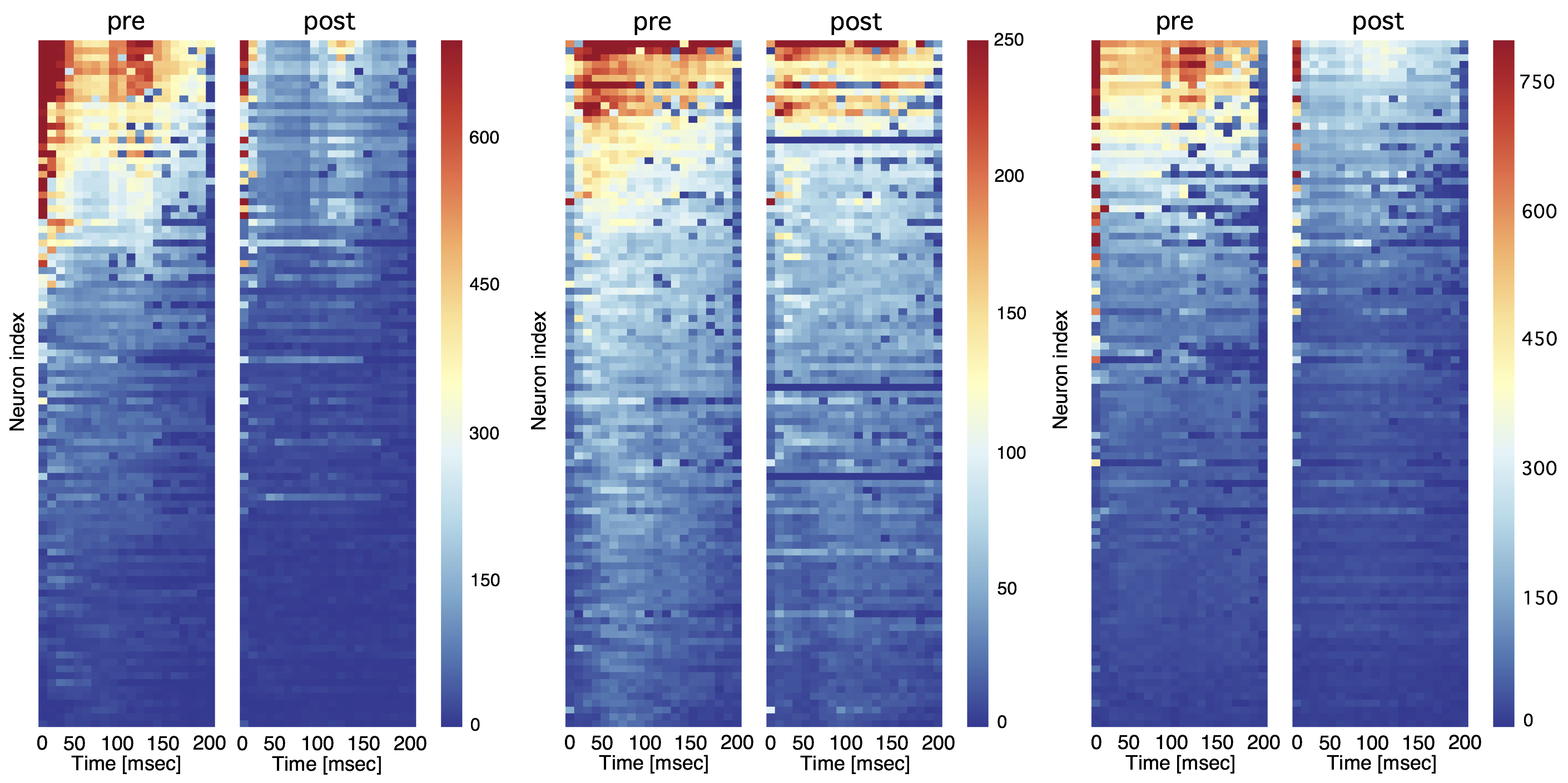}
\caption{Stimulus-evoked spikes for all neurons in failure cases from experiment 2; ``pre'' shows the first 5 min of the experiment, ``post'' shows the last 5 min. In all examples, the evoked spikes at the end of the experiment (i.e., accordint to ``post'') decreased.}
\label{fig:evoked_sample_2d}
\end{figure}

As shown in Fig.~\ref{fig:evoked_sample_2d}, the stimulus-evoked spikes decreased in all failure cases. These results are similar to the results from the open-loop conditions in experiment 1. 

\begin{figure}[htbp]
\centering
\includegraphics[width=12cm]{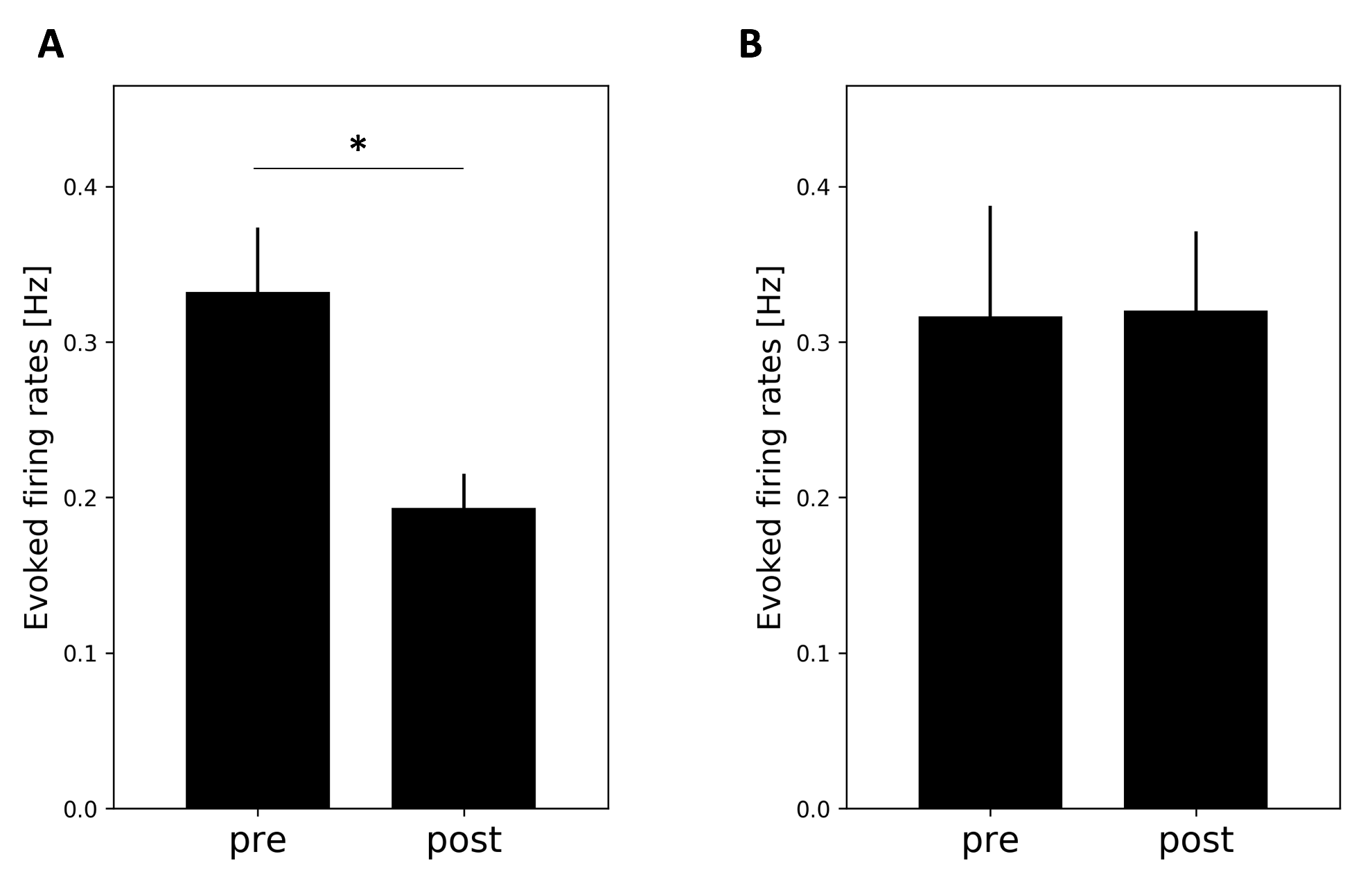}
\caption{Mean evoked firing rates with standard errors in the failure cases (n~=~3) from experiment 2. ``pre'' shows the first 5 min of the experiment, and ``post'' shows the last 5 min. A: mean evoked firing rate of all neurons, except for input neurons ($\text p<0.03$). B: mean evoked firing rates of input neurons.}
\label{fig:evoked_stats_robot}
\end{figure}

The statistical results showed that mean evoked firing rate of all neurons, except for input neurons, significantly decreased in the last 5 min of the experiment relative to the first 5 min (Wilcoxon signed-rank test, n~=~3, p~=~0.023) (Fig.~\ref{fig:evoked_stats_robot}A). On the other hand, the mean evoked firing rate of the input neurons did not change significantly. (Fig.~\ref{fig:evoked_stats_robot}B).

\begin{figure}[htbp]
\centering
\includegraphics[width=12cm]{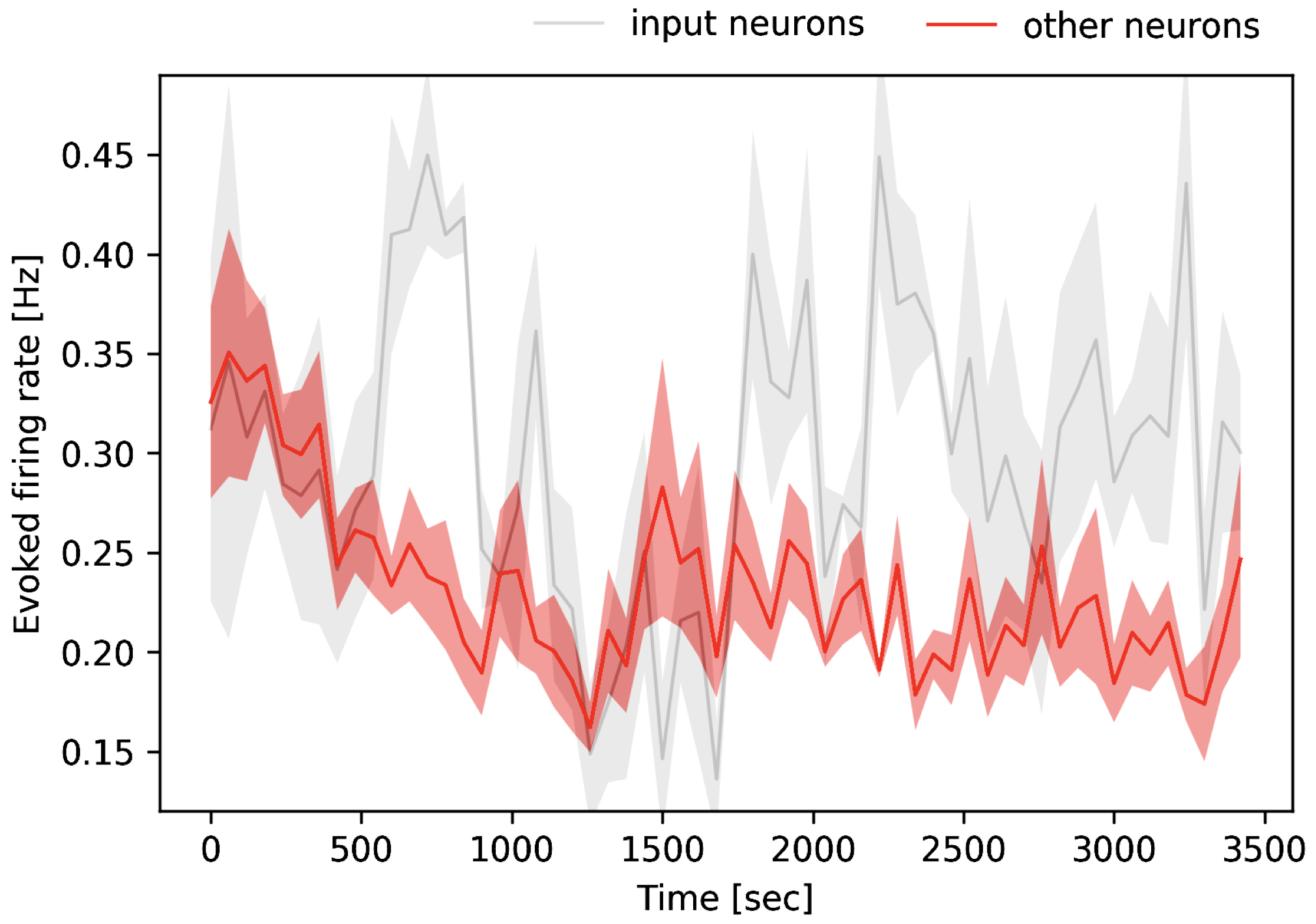}
\caption{Time series of evoked firing rates in the failure cases from experiment 2. The values represent the mean firing rate with standard errors in the specific time window after delivering a stimulus (50~ms for input neurons and 200~ms for other neurons).}
\label{fig:evoked_robot}
\end{figure}

Figure~\ref{fig:evoked_robot} shows the time evolution of the evoked firing rates of input neurons as well as other neurons. As shown in Fig.~\ref{fig:evoked_robot}, the evoked firing rates of other neurons gradually decreased during the experiments, but the evoked firing rates of input neurons did not decrease. 

These results imply that the decreased firing rates for all neurons, except for input neurons, was not caused by a decrease in the evoked spikes from the input neurons, but the weakening of the synaptic connection from the input neurons to the others. Therefore, this suggests that embodied cultured neural networks try to learn an action to avoid external stimulation, but, if the synaptic connections that control the motor output cannot stop the stimulus, the networks tend to ignore the external stimulation, thereby weakening the connection strength from the inputs. 


According to the results of the experiments, we found that if the cultured neurons cannot learn a behavior that avoids stimulation, the neural dynamics work to isolate the uncontrollable neurons which receive stimuli that the network cannot learn an action to avoid.

\subsection{Spiking Neural Networks Reproduce Stimulus-Avoiding Behaviors}
In the simulation experiments, we examined whether simulated networks reproduce the stimulus-avoiding behaviors observed in the experiments above, that is, when the network cannot learn an action to avoid input stimuli, the network tends to ignore the uncontrollable input by weakening the connections from the uncontrollable input neurons. 

\begin{figure}[htbp]
\centering
\includegraphics[width=12cm]{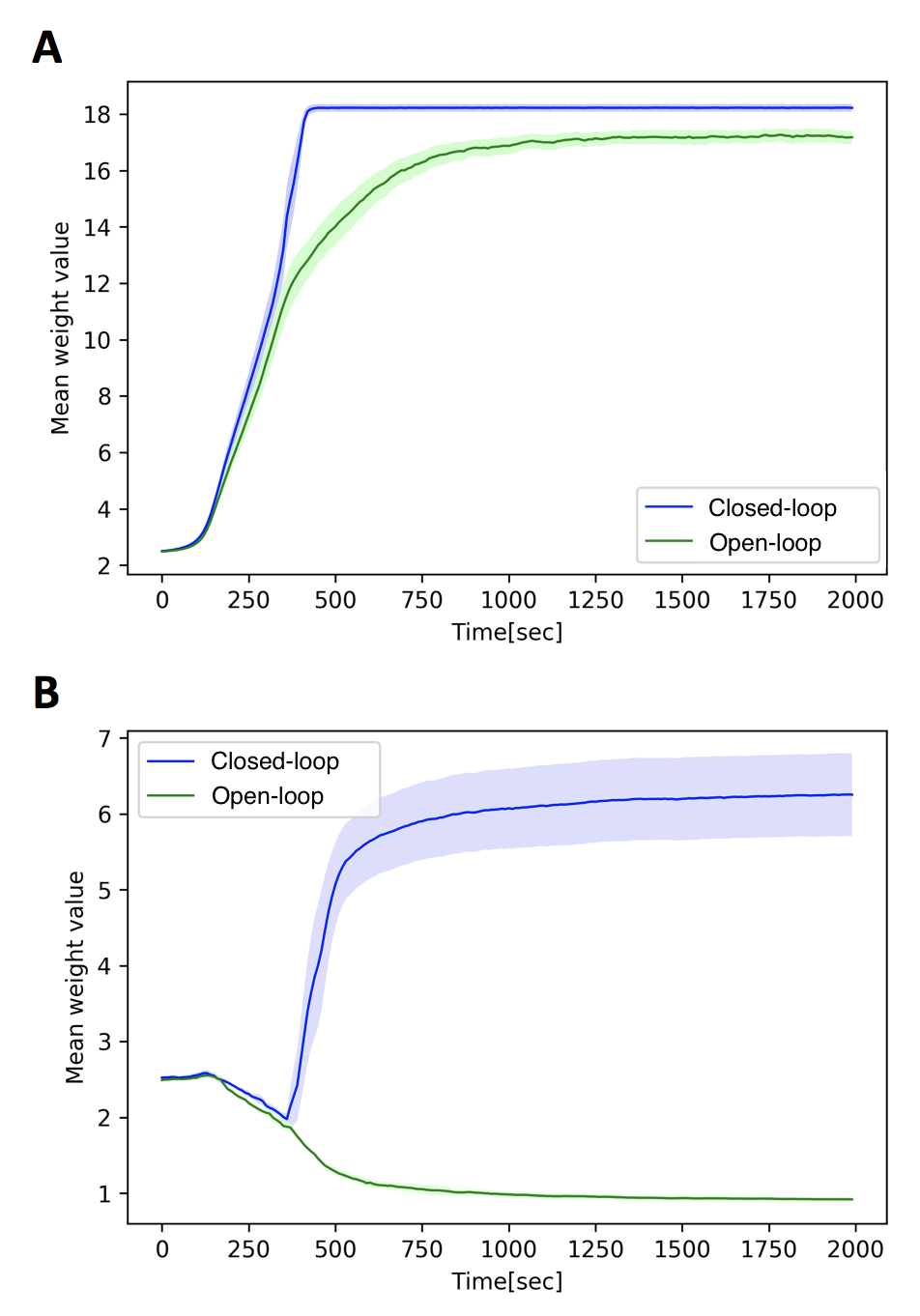}
\caption{Time evolution of the mean connection strength from the input neurons to the other neurons with standard error. A: Symmetric STDP (n~=~20). B: Asymmetric STDP: $A_{LTP}$~=~1.0, $A_{LTD}$~=~1.1; $\tau_{LTP}$~=~20, $\tau_{LTD}$~=~24 (n~=~20).}
\label{fig:weight_change}
\end{figure}

When applying symmetric STDP where the dynamics of LTP and LTD are symmetric ($A_{LTP}$, $A_{LTD}=1.0$; $\tau_{LTP}$, $\tau_{LTD}=20$), the synaptic weight from the input neurons increased in both closed-loop and open-loop conditions (Fig.~\ref{fig:weight_change}A). 

In the open-loop condition, the stimulation was randomly removed; thus, the network cannot learn the behavior to avoid the stimulation. The results show no dynamics leading to the isolation of the uncontrollable neurons. 

On the other hand, when applying asymmetric STDP where the dynamics of LTP and LTD are asymmetric ($A_{LTP}$~=~1.0, $A_{LTD}$~=~1.1; $\tau_{LTP}$~=~20, $\tau_{LTD}$~=~24; the peak of LTD is higher and the tail of LTD is longer than that of LTP), the synaptic weights from the input neurons increased in the closed-loop conditions, but it decreased in the open-loop conditions (Fig.~\ref{fig:weight_change}B). This tendency is the same as that observed in the experiments using the neuronal cultures explained above.
Therefore, we found that spiking neural networks with asymmetric STDP reproduce the dynamics to isolate the uncontrollable input neurons observed in the experiments using cultured neurons.

\begin{figure}[ht]
\centering
\includegraphics[width=12cm]{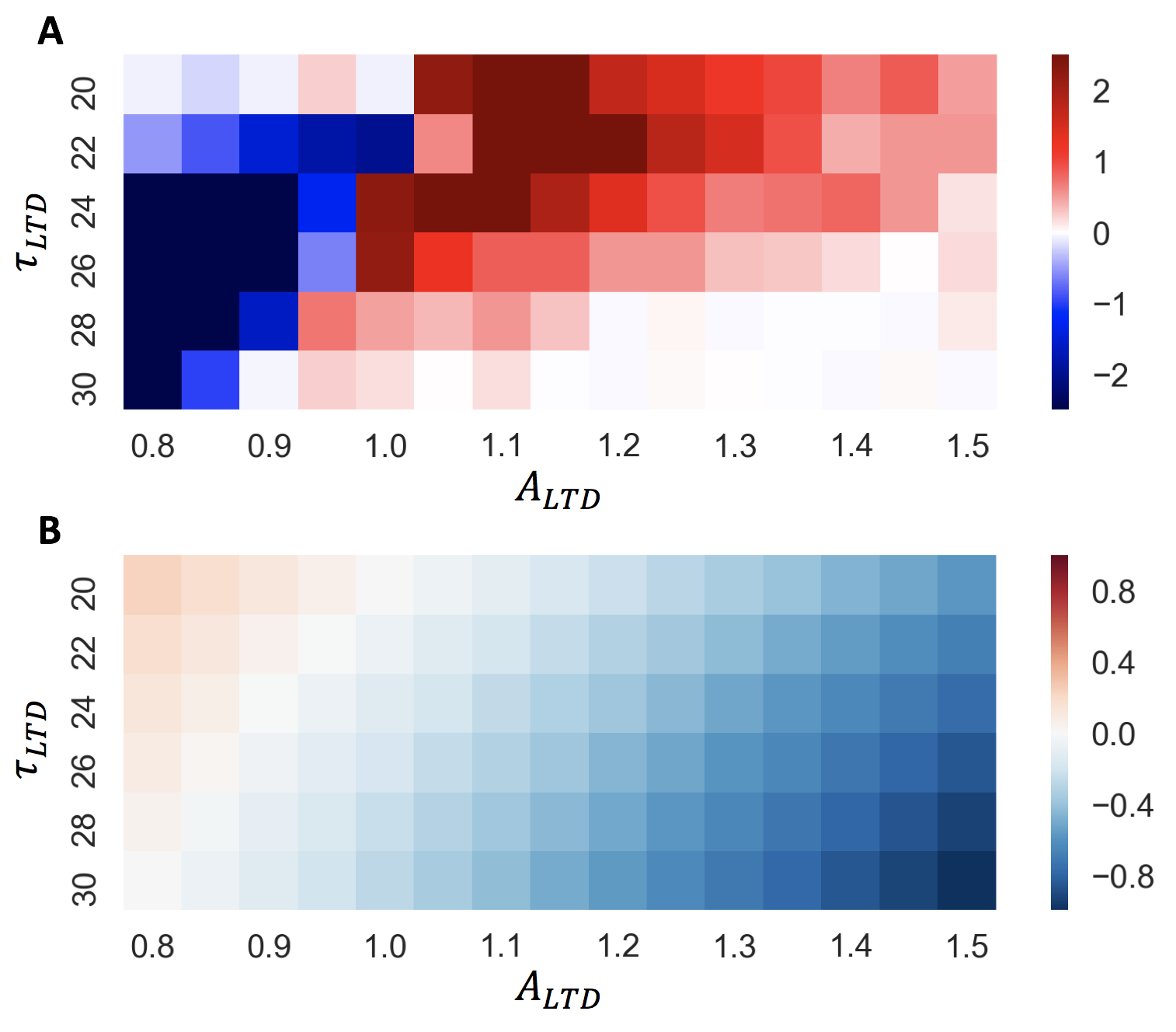}
\caption{A: Dependence of the performance of stimulus avoidance by weight selection on $A_{LTD}$ and $\tau_{LTD}$. The color represents the mean selection indicator ($SI$) with each parameters (n~=~20). B: Dependency of the STDP curve on the parameters $A_{LTD}$ and $\tau_{LTD}$. The color represents the integral value of STDP function with each parameters. }
\label{fig:SAS_parameter_search}
\end{figure}

To examine the conditions where this behavior works, we explored the parameter space-changing $A_{LTD}$ and $\tau_{LTD}$ in Eq.~\ref{eq:stdp}. $A_{LTD}$ is the parameter for the strength of LTD. $\tau_{LTD}$ represents the working time window of LTD.
In Fig.~\ref{fig:SAS_parameter_search}A, the color represents the value of the selection indicator ($SI$), which is defined as follows:

\begin{equation}
\label{eq:selection_indicator}
SI=W_{ci} - W_{oi}
\end{equation}

Here, $W_{ci}$ denotes the average weight of the connections from input neurons to other neurons in the closed-loop condition and $W_{oi}$ denotes that in the open-loop condition. $SI$ indicates the weight selection: a higher value indicates a higher selection tendency. The selection behavior here refers to the weights of the connections from input neurons with controllable inputs (i.e., those in the closed-loop condition) being reinforced, and weights from input neurons with uncontrollable inputs (i.e., those in the open-loop condition) being depressed.

The results of the asymmetric STDP with the parameters $A_{LTP}=1.0$, $A_{LTD}=1.1$; $\tau_{LTP}=20$, $\tau_{LTP}=24$ (Fig.~\ref{fig:stdp}B) show the maximum value of $SI$ in the parameter space. With $A_{LTP}=1.0$, $A_{LTD}=0.95$; $\tau_{LTP}=20$, $\tau_{LTP}=28$ (Fig.~\ref{fig:stdp}C), the shape of the STDP function is similar to that of the classical STDP function observed {\it in vitro} and {\it in vivo} \cite{Caporale2008}. The peak of LTD is lower than the peak of LTP, and the working time window of LTD is longer than that of LTP. The value of $SI$ in this region was still positive, implying the networks isolate the uncontrollable input neurons. This suggests that the dynamics can also work in biological neural networks. 

Figure~\ref{fig:SAS_parameter_search}B shows the integral values of the STDP function with the same parameters as in Fig.~\ref{fig:SAS_parameter_search}A .
In the blue regions, LTD is stronger than LTP; thus, random spikes in presynaptic neurons should decrease the weight from the neurons in theory. The weight selection occurred in the blue regions. However, if such a decrease was too strong (e.g., with the parameter: $A_{LTP}=1.0$, $A_{LTD}=1.4$; $\tau_{LTP}=20$, $\tau_{LTP}=30$; Fig.~\ref{fig:stdp}D), both $W_{ci}$ and $W_{oi}$ decreased to almost zero. As such, the weight selection did not occurred, because LTD was too strong compared to LTP; thus, almost all the weights of connections each neurons decreased and the networks cannot learn anything. Therefore, we found that the weight selection occurred in balanced regions where the integral value of LTD is stronger, but not much stronger, than that of LTP.

\begin{figure}[htbp]
\centering
\includegraphics[width=14cm]{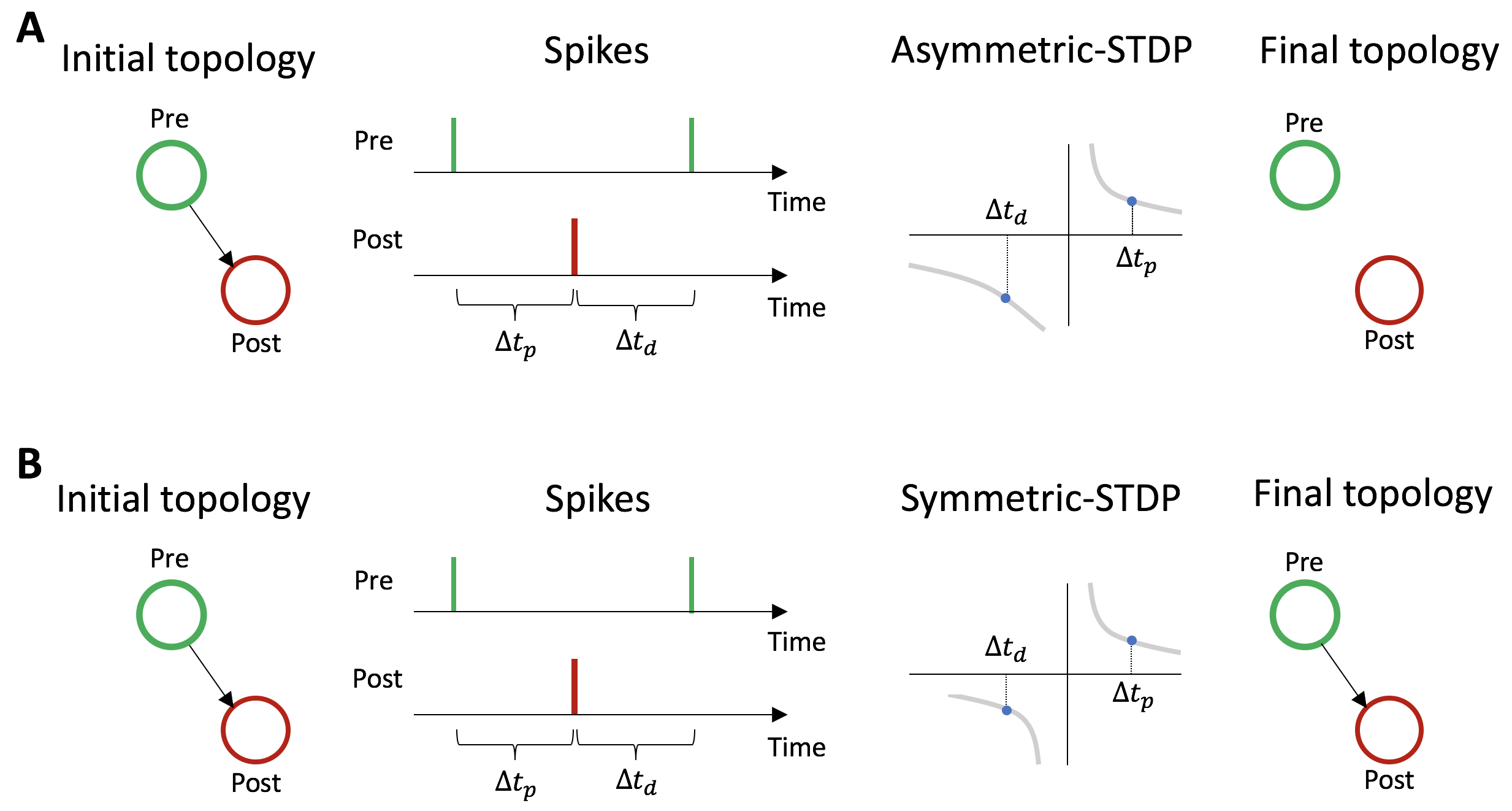}
\caption{Dynamics of the weight selection of 2 neurons: a presynaptic neuron, a postsynaptic neuron. A: Asymmetric STDP. If $\Delta t_{p} \approx \Delta t_{d}$ and $\Delta t_{p} < \tau_{LTP}$, $\Delta t_{d} < \tau_{LTD}$, the connection decreases because with asymmetric STDP, LTD has a greater effect than LTP ($\tau_{LTP}$ and $\tau_{LTD}$ are working time windows of LTP and LTD in the STDP function, respectively). B: Symmetric STDP. The connection between the neurons does not change much, even if the conditions above are satisfied, as LTP and LTD equally affect to the connection with the symmetric STDP.}
\label{fig:sas_dynamics}
\end{figure}

We examined what kind of connections were weakened by this weight selection dynamics in the spiking neural networks. To simplify, we considered a minimal case with 2 neurons and 1 connection (Fig.~\ref{fig:sas_dynamics}) to compare asymmetric STDP with symmetric STDP. In the asymmetric STDP case, if $\Delta t_{p} \approx \Delta t_{d}$ and $\Delta t_{p} < \tau_{LTP}$, $\Delta t_{d} < \tau_{LTD}$, the connection weight decreases because with the asymmetric STDP, LTD has stronger effect than LTP (Fig.~\ref{fig:sas_dynamics}A). In the symmetric STDP case, the connection does not change much because LTP and LTD affect the connection equally with the symmetric STDP (Fig.~\ref{fig:sas_dynamics}B). Thus, with the asymmetric STDP, if the mean value of the spike intervals that cause LTP ($\Delta tp$) and the mean value of the spike intervals that cause LTD ($\Delta td$) are close, as in $\sum_{t=1}^{N} \frac{\Delta tp_{i}}{N} \approx \sum_{t=0}^{N} \frac{\Delta td_{i}}{N}$, the connection disappears. 

In large networks, such situations occur if a presynaptic neuron fires independently from the other neurons at a high frequency. The connections from the input neurons, which are stimulated at a high frequency (e.g., more than 20~Hz when $\tau_{LTP}+\tau_{LTD} = 50$~msec) should decrease based on the dynamics of the asymmetric STDP. The stimulation frequency $SF$ [Hz] for the weight selection is as follows: $SF \geq k$, where $k \approx 10^3/(\tau_{LTP} + \tau_{LTD})$ for a minimal case with 2 neurons. This is inconsistent with our results regarding neuronal cultures, in which the networks showed a weight selection behavior at a low frequency (1~Hz). This suggests that this condition for stimulation frequency should be modified for larger networks.

In our experiments, we focused on the weight selection for input neurons; however, these dynamics should also work inside the networks. A synaptic weight from an active neuron with a high firing rate should decrease, as the conditions are consistent with that of input neurons with the high frequency stimulation explained above. However, in the case of hidden neurons, since there should be a feedback loop, this decrease leads to a decrease in the firing rate of the active neuron itself; if the firing rate becomes lower than a certain threshold, this weight depression should end. Thus, the synaptic weight should decrease but remain over zero, as in the case of the input neurons observed in our simulation experiments. This effect should stabilize the network through pruning connections from neurons with a high firing rate. 

\section{Discussion}
Our previous studies showed that embodied spiking neural networks with STDP learn a behavior as if they avoid stimulation, and the learning dynamics of the stimulus-avoiding behaviors scales from small networks to large networks (from two cells \cite{Sinapayen2016} to approx. 50,000 cells \cite{Masumori2018}). If LSA also works in cultured neurons, the smaller number of cultures than obtained by \cite{Shahaf2001} can learn an action to avoid stimulation. We found that a small number of cultured neurons (42--110 cells) can learn an action to avoid stimulation. This suggests that learning dynamics are based on LSA. In addition, we found that if the network could not learn the behavior to avoid the stimulus, plasticity works to suppress the influence of the uncontrollable stimulus on the network by weakening the connection from the input neurons. We also demonstrated that spiking neural networks with asymmetric STDP reproduced the stimulus-avoiding behaviors observed in the cultured neurons. Below, we further discuss the stimulus-avoiding behaviors.

In this study, we found that if a network cannot learn the behavior needed to avoid a stimulus, plasticity works to suppress the influence of an uncontrollable stimulus on the network by weakening the connection strength from the input neurons. In neuroscience, this kind of phenomena in which constant sensor inputs are ignored is known as neural adaptation, sensory adaptation, or stimulus-specific adaptation. These phenomena are observed in many regions in the brain (e.g., in the auditory system \cite{Ulanovsky2003, Shiramatsu2013, Noda2014, Parras2017}) and also {\it in vitro} \cite{Whitmire2016}. Such adaptation can be divided into two types: fast adaptation (less than one second) and slow adaptation (more than a few minutes) \cite{Chung2002}. The mechanism of slow adaptation is considered to be synaptic plasticity, as with LTD. While one of the possible mechanism of fast adaptation is synaptic fatigue \cite{Simons-Weidenmaier2006}, in which repeated stimulation depletes the neurotransmitters in the synapse, weakening the response of neurons to the stimulation. Some studies suggest that synaptic fatigue is caused by high-frequency stimulation, and occurs in presynaptic neurons \cite{Simons-Weidenmaier2006}. 

In our experiment on neuronal cultures, low-frequency stimulation (1~Hz) was used, and the results showed that the evoked spikes of the presynaptic neuron (input neurons) did not decrease, and the evoked spikes of all the other neurons decreased for more than 20 minutes. Therefore, our results with neuronal cultures suggest that the observed behavior resembles slow adaptation caused by LTD. The spiking neural networks with asymmetric STDP reproduced the observed slow adaptation (although the networks have STP that has similar dynamics to that of synaptic fatigue, STP stabilizes the firing rate within 1~sec; thus, slow adaptation cannot be caused by STP). Moreover, we found that the phenomena observed in parameter spaces where the shape of STDP function was similar to the shape broadly observed {\it in vitro}. Therefore, we argue that the mechanism based on asymmetric STDP should work in biological neural networks.

In the field of artificial life, adaptive behavior is and has been a major theme and should be explored further (e.g. see \cite{Paolo2000, Iizuka2007, DiPaolo2008}). Many researches have studied the emergence of adaptive behavior as an outcome of neural homeostasis. Although their models are very insightful, they are too abstract to explain homeostatic property in biological neural networks. In our previous studies, we found that bio-inspired spiking neural networks with STDP have homeostatic properties that allow networks to learn a behavior to avoid external stimulation from the environment (e.g., wall avoidance behavior) \cite{Sinapayen2016}. Our previous and present studies also showed that LSA can work in biological neural networks {\it in vitro} \cite{Masumori2015, Masumori2018} suggesting that LSA can explain homeostatic adaptation {\it in vivo}. 

Recently, Friston proposed the free energy principle (FEP) \cite{Friston2006, Friston2010} based on Bayesian inference by extending the predictive coding model \cite{Rao1999}. To minimize free energy (i.e., surprise) in FEP, the way of reconfiguring the internal model is called perceptual inference; on the other hand, reducing surprises using actions is called active inference. Active inference has attracted much attention from the viewpoint of autonomous behavior. Indeed, Friston discussed the relationship between the homeostatic adaptive behavior of animals and active inference \cite{Friston2011, Friston2012, Friston2016}. In our framework of stimulus avoidance in neural networks, regarding external stimulation as a surprise, the neural dynamics used to learn actions in order to avoid surprise is similar to an intuitive interpretation of active inference, where an agent behaves to minimize surprise. To examine this insight, we need to calculate the free energy in our framework in future research.

In addition, in this paper, we proposed a new principle, saying that if embodied neural networks cannot learn actions to avoid a stimulation, the networks can work to isolate the neurons that are receiving uncontrollable stimulation from the environment.
This two-layered homeostatic principle is similar to Ashby's theory of ultrastability, in which the system has two types of homeostasis; if the first regular homeostasis is unstable and its essential variables exceed the limits then the second homeostasis works to rearrange the system dramatically \cite{Ashby1960}. The system will reconstruct itself by trial and error until a stable homeostasis is acquired. Ashby suggests that biological systems are ultrastable with these two types of homeostasis \cite{Ashby1960}. Our results suggest that such a behavior can emerge thanks to the local dynamics of neurons in both biological and bio-inspired artificial neural networks.　

The simulation results showed that ignoring an uncontrollable constant stimulus is a strong feature of spiking neural networks with asymmetric STDP. Almost all the connection from the input neurons with uncontrollable stimuli decreased to zero. However, the synaptic weights from the input neurons with controllable inputs (i.e., a sensor input that the agent can learn to avoid) increased. This suggests that the networks can isolate the input neurons with uncontrollable stimulus inputs, and it can be regarded as dynamics trying to regulate self and non-self. A closed-loop of the sensor and motor, in which the motor outputs control the sensor stimulation like sensorimotor contingency \cite{ORegan2001} is regarded as self, while an open-loop of the sensor and motor is regarded as non-self. The open-loop collapses after isolating the sensor neuron. Thus the self-boundary is not limited to the network, but extended to the environment through its body.
It is interesting that the dynamics emerge from just the simple local dynamics of neurons.


How to discriminate self from non-self reminds us a theory of autopoiesis proposed by Maturana and Varela \cite{Maturana1980}. In addition to the structural viewpoint of regulating self-boundary explained above, here we discuss the results of self-regulating behavior from an autopoietic point of view. 

In autopoiesis, discrimination comes with the boundary between self and non-self. It is not a physical rigid boundary but a dynamic one: it should be constantly produced and maintained by its system's own processes.

Varela reported a simple mathematical model using artificial chemistry featuring autopoiesis \cite{Varela1974}. Two metabolite particles (S) generate one boundary particle (L) catalyzed by a catalytic particle (C). Those boundary particles connect to form a connected boundary, which encloses C and L. The boundary constantly decays and is repaired by the free boundary particles L. This self-organizing process of encapsulating C and L defines self-discrimination. No single particle defines the self-boundary; rather self-entity only emerges at a certain collective level.

This picture becomes much clearer by taking the immune systems as an example. Vertebrates establish self/non-self discrimination by forming an idiotype network, in which an antibody-antigen chain reaction exists among antibodies according to Jerne's hypothesis \cite{Jerne1974}. The current understanding of self/non-self discrimination has a molecular biological basis; however, the acquired immunity still needs to be exploited. A candidate for the explanation is the autopoietic picture. Each antibody can adaptively change the self-boundary. By self-organizing an idiotype network, self/non-self discrimination emerges as a result of the network reactions: the antigen-antibody reaction is suppressed locally for the self-antigens, but the reaction is percolated for the non-self antigens. The reaction network determines self/non-self discrimination similar to how Varela's simple artificial chemistry \cite{Varela1974} is determined.

Coming back to the present study, each neural responses do not determine the self/non-self boundary alone. Self/non-self boundaries emerge only in neural networks of a certain size. A neural network determines the self/non-self in the same way the immune system does. The boundaries of self/non-self for immune systems and for neural systems are processed dynamically. It is not explicit in the network whether a certain firing pattern of neuron depends on what comes from the outside or from the inside of the network. As in the immune system, a pattern which makes the network's response stronger and causes structural changes in the network is regarded as non-self here. 

For example, the controllable input above is initially regarded as a pattern from the outside of the network (non-self). However, as the change of the network progresses and the network learns the behavior to control it, the input will no longer cause large changes in the network and will be regarded as a pattern from the inside of the network (self). The uncontrollable input is initially regarded as a pattern from the outside (non-self) as in the case of controllable inputs; however, by weakening the connections from the sensor based on the dynamics proposed in this paper, explicit boundaries like Varela's cellular boundary \cite{Varela1974} can be created. As such, the inputs are explicitly isolated from the inside and no longer affects the internal network (self). 

Further, our previous results with spiking neural networks consisting of 100 neurons with almost the same models in this paper could predict a simple causal sequence of stimuli \cite{Masumori2018}. In that case, like the controllable input, predictable input is initially regarded as an external pattern (non-self) that causes structural changes in the network; however, when the network learns to predict the input, the input no longer affects the network and is regarded as a pattern from the inside (self). 

Although we have only discussed the patterns from environment, this dynamics should also work inside the network (although regulation by action does not occur inside the network and requires coupling with the environment). Inside the network, the dynamics of isolation of uncontrollable patterns and prediction of predictable patterns regulate the boundaries, and the network converges to stable states in which the network shows transitions of several patterns. In this way, the neural network can also be regarded as a system that acquires its own stability through the autopoietic process by means of action, prediction, and selection.

\section{Conclusion}
We presented neural homeodynamic responses to environmental changes based on experiments using both biological neural networks and artificial neural networks. As a result of the experiments, we found that the embodied neural networks show two kinds of stimulus-avoiding behaviors: (i) when input stimuli are controllable via actions, the embodied networks learn an action to avoid the stimulation and (ii) when input stimuli are uncontrollable, the connections from neurons with the uncontrollable input are weakened to avoid influences of the stimulation on other neurons. We argued that these stimulus-avoiding behaviors are regarded as dynamics of an autonomous regulation of self and non-self, in which controllable neurons are regarded as self, and uncontrollable neurons are regarded as non-self. This paper introduced neural autopoiesis by proposing the principle of stimulus avoidance. We thus extended the notion of autopoiesis to neural networks.

\section{Acknowledgments}
The high-density CMOS array used in this study was provided by Professor Andreas Hierlemann, ETH Z\"{u}rich. This work was supported by Grant-in-Aid for JSPS Fellows (16J09357), KAKENHI (17K20090), AMED (JP18dm0307009), the Asahi Glass Foundation, and the Kayamori Foundation of Information and Science Advancement. This work is partially supported by MEXT project ``Studying a Brain Model based on Self-Simulation and Homeostasis'' in
Grant-in-Aid for Scientific Research on Innovative Areas
``Correspondence and Fusion of Artificial Intelligence and Brain Science''
(19H04979)

\footnotesize
\bibliographystyle{apalike}
\bibliography{ALIFE2018_SAP_2} 

\end{document}